\documentclass{article}

\usepackage[preprint]{neurips_2026}

\usepackage[utf8]{inputenc} 
\usepackage[T1]{fontenc}    
\usepackage[pagebackref,breaklinks,colorlinks,citecolor=blue,linkcolor=blue,urlcolor=blue]{hyperref}
\usepackage{url}            
\usepackage{booktabs}       
\usepackage{amsmath}        
\usepackage{amsfonts}       
\usepackage{nicefrac}       
\usepackage{microtype}      
\usepackage{xcolor}         
\usepackage{graphicx}       
\usepackage{subcaption}

\title{Geometric Asymmetry in MoE Specialization: Functional Decorrelation and Representational Overlap}

\author{%
  Feilong Liu \\
  \texttt{drbruceliu@gmail.com} \\
  \texttt{https://www.linkedin.com/in/feilong-liu-19b6b18/} \\
}

\begin{document}

\maketitle

\begin{abstract}
Mixture-of-Experts (MoE) architectures achieve scalable capacity through sparse routing, yet the geometric structure of expert specialization 
remains poorly understood. We introduce a unified Jacobian-PCA-Grassmann framework for analyzing MoE layers in both function space and 
representation space. Across pretrained MoE Transformers (Mistral, Qwen), we find a consistent structural asymmetry: 
experts exhibit strong functional decorrelation—consistently low (near-zero) cross-expert Jacobian alignment—while their 
routed representations occupy distinct but partially overlapping subspaces. This indicates that functional decorrelation 
and representation overlap coexist rather than coincide in MoE specialization. Controlled routing experiments further 
indicate that routing sparsity appears to be a key factor shaping this geometry: Top-k routing induces sharper functional separation 
and larger subspace divergence, whereas fully-soft routing yields more entangled expert structure. Together, 
these results suggest a geometric interpretation in which MoE layers may be viewed as implementing locally decorrelated operators over overlapping submanifolds 
on a shared representation manifold, and provide a general diagnostic framework for studying conditional computation 
in modern Transformer architectures.
\end{abstract}

\section{Introduction}

Mixture-of-Experts (MoE) architectures have become a central mechanism for scaling Transformer models, enabling large 
parameter counts through sparse routing while maintaining fixed computational cost 
\citep{Shazeer2017SparselyGated, Lepikhin2020GShard, Fedus2022Switch}. Despite their widespread adoption in modern language models, 
the geometric structure of expert specialization remains poorly understood. Existing analyses primarily 
focus on routing statistics, load balancing, or system-level behavior \citep{Lepikhin2020GShard, Zoph2022STMoE, Eo2025MoCE}, 
leaving open fundamental questions about how experts differ in function space and how routed tokens organize in representation space.

While routing statistics provide a coarse view of expert behavior, a common intuition is that MoE layers partition 
the input distribution, with each expert acting as a local function approximator \citep{Jacobs1991AdaptiveMixtures, Jordan1994HME}. 
However, empirical studies have shown that experts often operate on overlapping token regimes 
\citep{Krishnamurthy2023ImprovingExperts, Dai2024DeepSeekMoE}, and that specialization does not always correspond to 
clean input-space clusters. This raises a deeper question: 
what geometric structure does routing induce inside a pretrained MoE Transformer?

To move beyond input-space intuitions, we study MoE specialization through a unified geometric lens \footnote{Code and data: \url{https://drive.google.com/drive/folders/1X1nP8A2s62JrVu8Pb_Bgrw9cvP776iSl?usp=sharing}}. 
We introduce a Jacobian-PCA-Grassmann framework that jointly analyzes (i) functional geometry, via expert-local Jacobians 
(following prior Jacobian-based analyses of neural sensitivity; e.g., \citep{Novak2018Sensitivity, Pennington2018Spectral}), 
and (ii) representation geometry, via routed PCA and subspace distances (building on representation-space analyses in Transformers;
 \citep{Ethayarajh2019Contextual, Timkey2021RogueDimensions}). 
 Applying this framework to pretrained MoE Transformers—including Mistral and Qwen—we uncover a consistent structural asymmetry: 
 experts exhibit strong functional decorrelation (near-zero cross-expert Jacobian alignment) while their routed representations 
 occupy distinct but partially overlapping subspaces. This suggests that functional decorrelation and representation overlap coexist 
 rather than coincide, providing evidence that is inconsistent with the assumption that MoE layers implement disjoint partitions of the input space.

To probe the mechanism underlying this asymmetry, we conduct controlled routing experiments and find that routing sparsity 
appears to be a key factor shaping the MoE geometry. Top-$k$ routing induces sharper functional separation and larger Grassmannian distances, 
whereas fully-soft routing yields more entangled expert geometry. These results indicate that routing sharpness modulates 
the degree of specialization, providing a mechanistic explanation for expert diversity in conditional computation 
\citep{Shazeer2017SparselyGated, Fedus2022Switch}.

Taken together, these findings support a view of MoE layers as implementing a soft partitioning \footnote{We use ``soft partitioning'' informally to denote 
overlapping but distinct submanifolds, not a strict mathematical partition.} of a shared representation 
manifold, where experts act as locally decorrelated operators over overlapping regimes. Beyond characterizing 
pretrained models, our framework provides a general diagnostic tool for analyzing expert specialization in conditional 
computation architectures.

\paragraph{Why this matters for scaling.} Modern large-scale MoE models—including Mixtral, DeepSeek-MoE (\citep{Dai2024DeepSeekMoE}), and Qwen-MoE—derive 
their efficiency from sparse routing, yet their internal specialization remains poorly understood. 
As these architectures scale to hundreds of experts and trillions of parameters, understanding how experts differentiate, 
when they collapse, and what geometric structure routing induces becomes critical for both training stability and 
inference reliability. Our results indicate that expert specialization is governed by a consistent geometric pattern—functional 
decorrelation paired with overlapping representation subspaces—and that routing sharpness directly controls this structure. 
This provides actionable insight for designing, scaling, and debugging MoE systems at frontier model sizes.

\subsection{Contributions}

This paper makes the following contributions:
\begin{itemize}
    \item \textbf{A unified geometric diagnostic framework for analyzing MoE specialization:}
    We introduce a unified Jacobian—PCA—Grassmann framework to jointly characterize MoE specialization in function space 
    (cross-expert Jacobian alignment) and representation space (routing-induced subspace structure via PCA and Grassmannian distance). 
    This provides a systematic diagnostic for studying expert specialization beyond routing statistics or activation patterns. 
    \item \textbf{Evidence of strong functional decorrelation across experts:} 
    Across pretrained MoE Transformers (Mistral, Qwen), we show that cross—expert Jacobian alignment is consistently near zero, 
    indicating that experts implement locally distinct transformations rather than redundant variants of the same function.
 
    \item \textbf{Discovery of overlapping but distinct representation subspaces:} 
    Despite functional decorrelation, routed representations occupy distinct yet partially overlapping subspaces, 
    as quantified by Grassmannian distances. This suggests a structural asymmetry,
    where MoE layers appear to separate functions more strongly than input clusters.
  
    \item \textbf{A geometric interpretation of MoE layers as soft partitioning of a shared manifold:} 
  These findings support a geometric interpretation of MoE layers as soft partitioning of a shared manifold, where experts act as 
  locally decorrelated operators over overlapping regimes.

\end{itemize}

\subsection{Relation to prior work}

\textbf{Mixture—of—Experts routing and scaling:}
Early MoE work established sparse routing as an effective strategy for scaling neural networks 
\citep{Shazeer2017SparselyGated, Lepikhin2020GShard, Fedus2022Switch}. Subsequent studies refined routing mechanisms, 
load-balancing objectives, and training stability \citep{Eo2025MoCE,Krajewski2024ScalingLaws}. 
These works primarily analyze routing behavior and system-level performance, but do not characterize 
the geometric structure of expert specialization inside pretrained MoE Transformers.

\textbf{Expert specialization and redundancy:}
Beyond routing behavior, several works examine expert redundancy, merging, and pruning 
\citep{Li2025SubMoE, Miao2025MergeMoE, Yang2024MoEI2, Huang2025MixtureCompressor}. These studies suggest that experts 
may be partially redundant, but rely on performance-based or parameter-based criteria. 
They do not analyze functional geometry (Jacobian structure) or representation subspaces, 
leaving an open question: when do experts represent genuinely distinct functions, 
and when do they reflect redundant parameterization?

\textbf{Jacobian analysis in deep networks:}
Orthogonal to MoE—specific studies, Jacobian—based analyses have been used to study sensitivity, curvature, 
and robustness in dense networks \citep{Dadoun2025JacobianStability}. However, these analyses focus on monolithic architectures 
and do not address conditional computation. In particular, prior work does not examine cross‑expert Jacobian 
alignment or functional decorrelation in MoE layers.

\textbf{Representation geometry and subspace structure:}
Complementary to functional analyses, representation-space studies examine representation anisotropy and layer geometry 
in Transformers \citep[e.g.,][]{Ethayarajh2019Contextual, Timkey2021RogueDimensions}. 
Grassmannian distances \citep{Edelman1998Orthogonality, Absil2008Manifolds, Golub2013MatrixComputations} 
have been used to compare learned subspaces in multitask or modular networks, but not in the 
context of MoE routing. To our knowledge, no prior work jointly analyzes Jacobian geometry and routed representation 
subspaces in pretrained MoE Transformers.

\textbf{Our contribution in context:}
Prior work on MoE models has primarily examined routing behavior, load balancing, or expert redundancy, but has not analyzed 
the geometric structure of expert specialization. Existing studies do not measure cross—expert Jacobian alignment, 
do not characterize routed representation subspaces, and do not connect routing sparsity to geometric separation.

Our work fills this gap by providing the joint analysis of functional geometry (via expert—local Jacobians) 
and representation geometry (via routed PCA and Grassmannian distances) in pretrained MoE Transformers. 
This positions our framework as a geometric and mechanistic complement to existing MoE research, 
rather than a scaling or performance study.

\section{Methodology}
We analyze expert specialization in MoE Transformers through a unified geometric framework that characterizes 
(i) functional geometry via expert-local Jacobians and (ii) representation geometry via routed PCA and 
Grassmannian subspace distances. This section formalizes the probes and describes how they are applied 
to pretrained MoE models.

\subsection{Expert-Local Jacobians}
For an MoE layer with routing weights $g(x) \in \mathbb{R}^E$ and a collection of $E$ experts $\{ f_e \}_{e=1}^E$,
the layer output is

\begin{equation}
f_{\text{MoE}}(x) = \sum_{e=1}^{E} g_e(x)\, f_e(x).
\end{equation}

We study \textbf{the expert-local Jacobian}
\begin{equation}
J_e(x) = \frac{\partial f_e(x)}{\partial x} \in \mathbb{R}^{d_{\text{out}} \times d_{\text{in}}}.
\end{equation}

which captures the local linear behavior of expert $e$ independent of routing. This isolates functional 
specialization from routing dynamics. Accordingly, our analysis characterizes the geometry of expert 
functions under fixed routing, rather than the full end-to-end MoE mapping.

\textbf{Cross-expert Jacobian alignment (function geometry)}

To quantify functional overlap between experts, we measure alignment between their average expert-local Jacobians.
Let  $\bar{J}_e$ be the average Jacobian for each expert over the input distribution:

\begin{equation}
\bar{J}_e = \mathbb{E}_{x \sim D}\!\left[ J_e(x) \right].
\end{equation}

In practice, this expectation is approximated by averaging over a dataset:

\begin{equation}
\bar{J}_e = \frac{1}{N} \sum_{i=1}^N J_e(x_i).
\end{equation}

We then vectorize each averaged Jacobian:

\begin{equation}
v_e = \operatorname{vect}(\bar{J}_e) \in \mathbb{R}^{d_{\text{out}} \cdot d_{\text{in}}}.
\end{equation}

The pairwise functional alignment between experts $e$ and $e'$ is measured via normalized Frobenius
inner products (cosine similarity):

\begin{equation}
\operatorname{sim}(e, e')
= \frac{\langle v_e, v_{e'}\rangle}
       {\|v_e\|_2 \, \|v_{e'}\|_2}.
\end{equation}

Low cosine similarity ($\operatorname{sim}(e, e') \approx 0$) indicates functional decorrelation.

\subsection{Routed representation geometry}
Given a dataset $\{x_i\}_{i=1}^{N}$ for each expert $e$, we collect hidden states $h(x) \in \mathbb{R}^d$ weighted by routing weights $g_e(x) \in [0,1]$. This yields a routed dataset

\begin{equation}
H_e =
\begin{bmatrix}
h_e(x_1)^{\top} \\
h_e(x_2)^{\top} \\
\vdots \\
h_e(x_N)^{\top}
\end{bmatrix}
\in \mathbb{R}^{N \times d}.
\end{equation}

where $h_e(x_i) = g_e(x_i)\, h(x_i) \in \mathbb{R}^d$.

\textbf{Routed PCA}

We perform standard PCA on $H_e$ to obtain principal directions and variance spectra 
(eigenvalues $\lambda_{1,e} \ge \lambda_{2,e} \ge \cdots \ge \lambda_{d,e} \ge 0$).  
The Explained variance is $v_i = \frac{\lambda_i}{\sum_{k=1}^{d} \lambda_k}$, and cumulative variance is  
$CV_k = \frac{\sum_{i=1}^{k} \lambda_i}{\sum_{i=1}^{d} \lambda_i}$.   
Variance spectra and variance concentration quantify representational specialization.

\textbf{Subspace comparison via Grassmannian (Riemannian) Geodesic Distance}

Let $Q_i,\, Q_j \in \mathbb{R}^{d \times n}$ denote the top-$n$ PCA principal subspaces for experts $e_i$ and $e_j$, 
their geometric separation is measured by the Grassmannian Geodesic Distance 
\citep{Edelman1998Orthogonality, Absil2008Manifolds, Golub2013MatrixComputations}:

\begin{equation}
d_G(e_i, e_j) = \left( \sum_{m=1}^{n} \theta_m^2 \right)^{1/2}.
\end{equation}

Where $\theta_m = \arccos(\sigma_m) \in [0, \tfrac{\pi}{2}]$, and $\sigma_1,\, \sigma_2,\, \ldots,\, \sigma_n$ are singular values 
of $M = Q_i^{\top} Q_j \in \mathbb{R}^{n \times n}$.  
Theoretically, the maximum distance when perfectly orthogonal is $\frac{\pi}{2}\,\sqrt{n}$. 
For $n=5$, that maximum is approx 3.51.

When $d_G(e_i, e_j) \approx 0$, the subspaces are nearly identical, indicating redundant functional roles.
When $d_G(e_i, e_j) \approx \frac{\pi}{2}\,\sqrt{n}$, the subspaces are nearly orthogonal, indicating low overlap in local linear function spaces.
Unlike cosine similarity—which compares individual directions—this metric captures multi-directional separation between entire subspaces, 
making it sensitive to higher-dimensional differences in local function structure. In addition, this metric is scale-invariant 
and depends only on subspace orientation, making it robust to differences in activation magnitude across models.

\subsection{Routing-Dependent Geometry}
To isolate the effect of routing, we evaluate the same controlled 3-layer Transformer model under 
(1) Top-k routing (sparse, discrete selection) and (2) Fully-soft routing (dense, continuous mixture)
while keeping expert parameters fixed. This allows us to attribute geometric differences directly to 
routing sparsity rather than training dynamics or parameter changes.

\subsection{Models and Evaluation Protocol}
We evaluate MoE geometry using both pretrained MoE models and controlled 3-Layer Transformer models. 
This unified protocol ensures consistent comparisons across experts, layers, datasets and routing regimes.

\textbf{Pretrained MoE models (Primary Analysis):}
We evaluate geometric structure on large-scale pretrained models to assess expert specialization in practical settings. 
Specifically, we compare dense and MoE variants of: (i) Mistral-7B (dense) vs Mistral-8{x}7B (MoE); 
and (ii) Qwen2.5-1.5B (dense) vs Qwen1.5-MoE-A2.7B (MoE).
These models allow us to examine how sparse routing affects functional and representation 
geometry in architectures used in modern language modeling.

\textbf{Controlled 3-Layer Transformer model (Routing Mechanism Isolation):}
To isolate routing effects independently of large-scale training dynamics, we conduct controlled 
experiments using a 3-layer Transformer model with 8-expert MoE layers, trained from scratch on WikiText. 
In this setting, we vary only the routing mechanism—Top-k versus fully-soft routing—while keeping 
all other architectural and training factors fixed. This controlled environment provides a clean 
testbed for attributing geometric differences directly to routing sparsity.

\textbf{Data and batch consistency:}
For pretrained models, we run inference over the full evaluation datasets. For the controlled model, 
we use the entire WikiText corpus for both training and evaluation.
For each expert, the Jacobian matrix, routed PCA and Grassmannian distances are computed on the same routed batches, 
ensuring direct comparability between functional and representation geometry.

\textbf{Protocol Summary:}
This evaluation protocol provides a consistent basis for comparing:
(i) dense vs. MoE architectures;
(ii) Top-k vs. fully-soft routing; and
(iii) pretrained vs. controlled settings,
and ensures that observed geometric patterns reflect expert specialization rather than sampling or evaluation artifacts.

\section{Results}
We evaluate dense and MoE models using the geometric probes introduced in Section 2: (i) expert-local Jacobians, 
which characterize first-order functional structure, and (ii) routed PCA and Grassmannian distances, which 
characterize representation geometry. We use Mistral for visualization; results are consistent across Qwen (Section 3.4).

\begin{figure}[t]
    \centering

    \begin{subfigure}{0.48\linewidth}
        \centering
        \includegraphics[width=\linewidth]{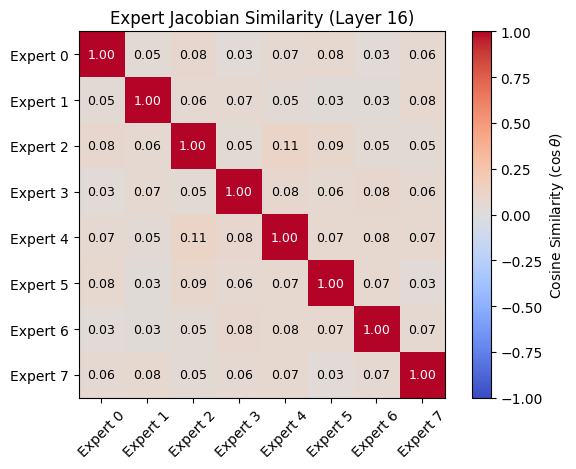}
        \caption{}
    \end{subfigure}
    \hfill
    \begin{subfigure}{0.48\linewidth}
        \centering
        \includegraphics[width=\linewidth]{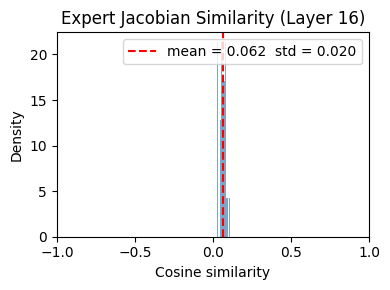}
        \caption{}
    \end{subfigure}

    \vspace{0.25em}

    \begin{subfigure}{0.48\linewidth}
        \centering
        \includegraphics[width=\linewidth]{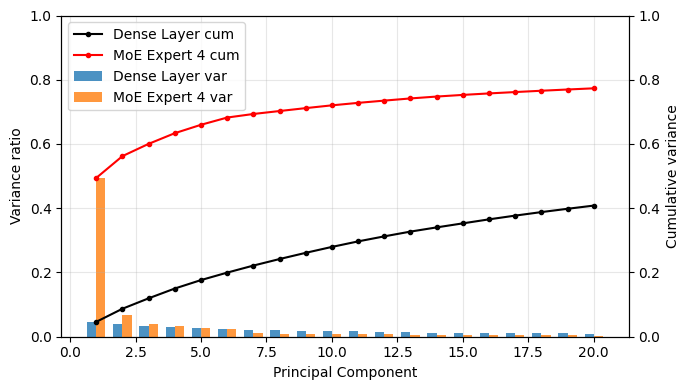}
        \caption{}
    \end{subfigure}
    \hfill
    \begin{subfigure}{0.48\linewidth}
        \centering
        \includegraphics[width=\linewidth]{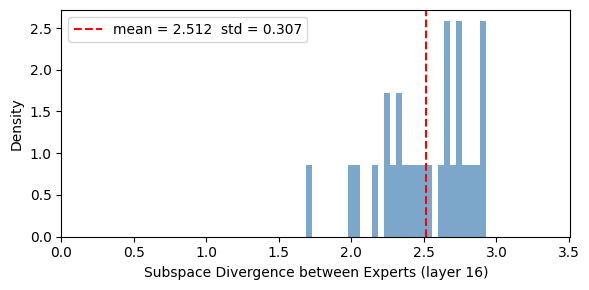}
        \caption{}
    \end{subfigure}

    \caption{\textbf{Geometric structure of MoE specialization in Mistral-8$\times$7B (Layer 16).}
    (a) Cross-expert Jacobian similarity matrix.
    (b) Distribution of Jacobian similarities.
    (c) Routed PCA spectra for dense vs.\ MoE expert layer.
    (d) Grassmannian distances between expert subspaces (top-5 components; theoretical maximum $\approx 3.51$).}
\end{figure}

\subsection{Functional decorrelation: Jacobian Geometry of Experts}
To assess whether MoE experts implement distinct local transformations, we measure cross‑expert Jacobian alignment. 
If experts were redundant, their Jacobians would exhibit high similarity.

\textbf{Cross-expert alignment}.  Figures 1a-b show the cosine similarity between expert-local Jacobians in Mistral (Layer 16). 
Off-diagonal similarities cluster tightly near zero (\(\mathrm{mean}=0.062,\ \sigma\approx0.020\)), 
indicating minimal shared sensitivity directions.

\textbf{Interpretation}.  Experts exhibit first-order functional decorrelation: each expert implements a 
locally distinct transformation rather than a scaled variant of its peers. This provides evidence that 
MoE specialization manifests as separation in function space.

\subsection{Representation Geometry: Subspace Structure of Routed Representations}
Jacobian decorrelation does not determine how expert representations are arranged in the model's hidden space. 
To quantify representational separation, we compute Grassmannian distances 
between the top-5 PCA subspaces of each expert.

\textbf{Subspace divergence}.  Figure 1d shows the distribution of pairwise Grassmann distances (\(\mathrm{mean}=2.512,\ \sigma=0.307\)). 
These values are large—indicating substantial separation—but remain below the theoretical maximum 
($\frac{\pi}{2}\,\sqrt{n} = \frac{\pi}{2}\,\sqrt{5} \approx3.51$ for complete orthogonality), 
implying non-orthogonal overlap.

\textbf{Interpretation}.  Experts occupy distinct but partially overlapping representation subspaces. 
This suggests that MoE layers may not fully partition the representation space into disjoint regions; 
instead, they form overlapping submanifolds with expert‑specific dominant directions.

\subsection{Coupled structure: functional decorrelation vs representation overlap}
Combining Sections 3.1 and 3.2 suggests a structural asymmetry:
(i) Function space: experts are strongly decorrelated (near-zero Jacobian alignment); and
(ii) Representation space: experts occupy distinct but overlapping subspaces (high Grassmann distances but remain below the theoretical maximum).

This indicates that functional decorrelation does not imply representational disjointness. 
Experts can be interpreted as acting as locally specialized operators applied to overlapping regimes on a shared manifold. 
We refer to this pattern as coupled geometric specialization.

\subsection{Cross-model consistency of MoE Geometry: Qwen and Mistral}
To assess whether these geometric patterns generalize across architectures, we repeat the analysis on Qwen-MoE.

\textbf{Functional geometry:}  Figures 2a-b (Appendix~D) show that Qwen's cross-expert Jacobian alignment is tightly centered at zero 
(\(\mathrm{mean}=0.000,\ \sigma=0.023\)), consistent with Mistral.

\textbf{Representation geometry:}  Grassmann distances (Fig. 2d, Appendix~D) exhibit a similar distribution 
(\(\mathrm{mean}=2.382,\ \sigma=0.379\)), indicating comparable subspace separation.

\textbf{Summary across models, layers, and datasets:}
To consolidate the geometric patterns observed across architectures, Table 1 reports the mean ± standard deviation 
of cross-expert Jacobian similarity and Grassmannian distance for Mistral and Qwen across early, middle, 
and late MoE layers and two evaluation datasets. The values show consistent low Jacobian alignment 
and substantial but sub-maximal subspace separation across all settings.

\begin{table}[t]
\centering
\small
\begin{tabular}{l c c c c}
\toprule
\textbf{Model} & \textbf{Layer} & \textbf{Dataset} &
\textbf{Jacobian Similarity} & \textbf{Grassmann Distance} \\
\midrule
Mistral-8$\times$7B & Early (3)  & 1 & $0.085 \pm 0.020$ & $2.061 \pm 0.303$ \\
Mistral-8$\times$7B & Middle (16) & 1 & $0.068 \pm 0.016$ & $2.537 \pm 0.324$ \\
Mistral-8$\times$7B & Late (29)  & 1 & $0.034 \pm 0.017$ & $2.484 \pm 0.311$ \\
Mistral-8$\times$7B & Early (3)  & 2 & $0.064 \pm 0.016$ & $2.126 \pm 0.225$ \\
Mistral-8$\times$7B & Middle (16) & 2 & $0.062 \pm 0.020$ & $2.512 \pm 0.307$ \\
Mistral-8$\times$7B & Late (29)  & 2 & $0.041 \pm 0.021$ & $2.547 \pm 0.280$ \\
\midrule
Qwen1.5-MoE-A2.7B & Early (3)  & 1 & $0.000 \pm 0.024$ & $2.499 \pm 0.283$ \\
Qwen1.5-MoE-A2.7B & Middle (16) & 1 & $0.001 \pm 0.024$ & $2.418 \pm 0.349$ \\
Qwen1.5-MoE-A2.7B & Late (22)  & 1 & $0.000 \pm 0.023$ & $2.689 \pm 0.261$ \\
Qwen1.5-MoE-A2.7B & Early (3)  & 2 & $0.001 \pm 0.024$ & $2.444 \pm 0.258$ \\
Qwen1.5-MoE-A2.7B & Middle (16) & 2 & $0.000 \pm 0.023$ & $2.382 \pm 0.379$ \\
Qwen1.5-MoE-A2.7B & Late (22)  & 2 & $0.000 \pm 0.023$ & $2.598 \pm 0.536$ \\
\bottomrule
\end{tabular}
\caption{\textbf{Summary of geometric metrics across models, layers, and datasets.} Mean ± standard deviation of 
cross-expert Jacobian similarity and Grassmannian distance (top-5 PCA subspaces).}
\end{table}

\textbf{Synthesis:} Both models exhibit the same geometric structure: low Jacobian alignment + high but 
sub-maximal Grassmann separation.  This suggests that the observed specialization pattern is a robust 
property of MoE architectures rather than an artifact of a specific model.

\subsection{Dense vs MoE: Differences in Representation Geometry}
We compare routed PCA spectra of MoE experts with dense feedforward layers.

\textbf{Variance spectra:}  Figure 1c\&2c shows that MoE experts exhibit steeper eigenvalue decay: the first four 
principal components capture $\sim 50\%$ of variance, compared to $< 20\%$ in the dense layer, for Mistral and Qwen respectively.

\textbf{Effective dimensionality:}  Figure 1c\&2c shows that MoE experts reach $\sim 80\%$ cumulative variance within the 
first 20 components, whereas dense layers require many more, for Mistral and Qwen respectively.

\textbf{Interpretation:}  Routing concentrates tokens into lower‑dimensional subspaces, yielding more focused 
representation structure than dense layers.

\subsection{Routing Mechanism: Top-k vs Fully-Soft Routing Effects on Geometry}
To isolate the effect of routing sharpness, we compare Top-k routing with fully-soft routing in the controlled 3-layer MoE model.

\textbf{Functional geometry:} Top-k routing preserves low Jacobian alignment (Fig. 3b\&d, Appendix~E), whereas fully-soft routing 
produces high-alignment peaks (Fig. 3a\&c, Appendix~E), indicating collapse toward redundant mappings.

\textbf{Representation geometry:}
Grassmann distances remain high under Top-$k$ routing (\(\mathrm{mean}=2.463\); Fig.~4d, Appendix~E) but collapse 
under fully-soft routing (\(\mathrm{mean}=0.480\); Fig.~4c, Appendix~E), indicating substantial subspace overlap.

\textbf{Interpretation:} 
Routing sparsity acts as a geometric separator:
(i) Top-k induces functional decorrelation and subspace separation; and 
(ii) Fully-soft routing collapses both functional and subspace structures.
These results provide evidence that routing sharpness plays a central role in shaping 
MoE geometric specialization.

\section{Discussion}
Our results reveal a consistent structural asymmetry in expert specialization across pretrained 
MoE Transformers: experts are functionally decorrelated—exhibiting near-zero cross-expert Jacobian 
alignment—while their routed representations remain distributed over distinct but non-orthogonal 
subspaces. This suggests that functional decorrelation and representational separation do not 
necessarily coincide. Rather than forming disjoint partitions of the input space, MoE layers organize computation into 
locally decorrelated transformations operating over partially shared representation regimes. 
In this sense, decorrelation serves as a geometric proxy for approximate functional independence, 
without implying full statistical independence.

The controlled routing experiments further clarify the mechanism behind this structure. 
Increasing routing sparsity (Top-k) sharpens functional decorrelation and enlarges inter-expert 
Grassmannian distances, while fully soft routing produces more entangled geometry. 
This identifies routing sharpness as a direct control knob for the strength of specialization, 
linking architectural design to measurable geometric structure.

These findings suggest that geometric probes—combining Jacobian alignment with subspace-based 
representation analysis—offer a principled way to characterize conditional computation beyond 
routing statistics alone. In particular, they expose a distinction between what experts compute 
(function space) and where they operate (representation space), which is not captured 
by standard activation-based analyses.

This distinction has potential implications for model design. For example, expert merging 
or pruning strategies based purely on representation similarity may incorrectly treat 
overlapping subspaces as redundancy, even when experts implement distinct functions. 
Our results suggest that functional metrics, such as Jacobian alignment, may provide a 
more reliable criterion for identifying true redundancy. We leave systematic 
validation of such design strategies to future work.

\section{Limitations}
While our framework provides a unified geometric view of MoE specialization, it is not intended as a complete 
account of MoE behavior. Our analysis focuses on pretrained models and emphasizes first-order functional 
structure (via Jacobians) and linear representation geometry (via routed PCA and Grassmannian distances). 
As a result, several important aspects remain outside the scope of this study.

First, the Jacobian-based analysis is local by construction and does not capture global nonlinear behavior 
or long-range trajectories in representation space. Similarly, the PCA-based representation analysis 
summarizes dominant subspace structure but does not fully characterize higher-order interactions between experts.

Second, our experiments are conducted on pretrained models and controlled routing variants, and therefore 
do not address how geometric structure evolves during training or under different optimization regimes, 
data distributions, or routing regularizers.

Finally, while our controlled experiments isolate the effect of routing sharpness in small models and 
show trends consistent with larger systems, further work is needed to validate these mechanisms at 
frontier scales, in deeper layers, and in more complex settings such as multimodal MoEs.

Extending geometric analysis along these directions would help clarify the generality and 
limitations of the patterns identified in this work.

\section{Future work}
Our geometric framework opens several directions for extending the analysis of conditional computation. 
One avenue is to study how functional and representation geometry evolve during training, 
particularly during phases where experts emerge, collapse, or reorganize. Another direction 
is to apply the framework to larger and more diverse MoE architectures—including multimodal models, 
hierarchical routing schemes, and architectures with hundreds of experts—to test whether the 
functional-representation asymmetry persists at frontier scales. Extending the analysis to nonlinear 
probes or higher-order geometric measures may also reveal additional structure beyond local 
Jacobians and linear subspaces.

A direct practical application lies in model compression: our finding that functional decorrelation coexists 
with representational overlap suggests that expert pruning or merging strategies based solely on activation 
similarity may incorrectly remove functionally distinct experts. Incorporating Jacobian alignment into 
redundancy metrics offers a principled alternative for identifying truly redundant experts without conflating 
overlapping subspaces with shared function. We leave systematic validation of such geometry-informed 
compression strategies to future work.

Finally, connecting geometric specialization to downstream 
behavior—such as robustness, calibration, or task-specific performance—offers a path toward 
mechanistic understanding of when and why MoE specialization is beneficial.

\section{Conclusion}
We introduced a unified Jacobian-PCA-Grassmann framework for analyzing the geometric 
structure of expert specialization in MoE Transformers. Across Mistral and Qwen, 
we identified a consistent structural asymmetry in expert specialization: experts 
are functionally decorrelated yet operate on overlapping representation subspaces. 
Controlled routing experiments further showed that routing sparsity modulates this geometry, 
with Top-k routing producing sharper separation than fully-soft routing. Together, 
these results support a view of MoE layers as implementing a soft partitioning of a shared 
representation manifold and provide a general diagnostic tool for studying conditional 
computation in modern Transformer architectures.

\bibliographystyle{unsrtnat}
\bibliography{references_jacobian}

@article{Fedus2022Switch,
  title={Switch Transformers: Scaling to Trillion-Parameter Models with Simple and Efficient Sparsity},
  author={Fedus, William and Zoph, Barret and Shazeer, Noam},
  journal={Journal of Machine Learning Research},
  year={2022}
}

@article{Shazeer2017SparselyGated,
  title={Outrageously Large Neural Networks: The Sparsely-Gated Mixture-of-Experts Layer},
  author={Shazeer, Noam and Mirhoseini, Azalia and Maziarz, Krzysztof and Davis, Andy and Le, Quoc and Hinton, Geoffrey and Dean, Jeff},
  journal={arXiv preprint arXiv:1701.06538},
  year={2017}
}

@inproceedings{Lepikhin2020GShard,
  title={GShard: Scaling Giant Models with Conditional Computation},
  author={Lepikhin, Dmitry and Lee, HyoukJoong and Xu, Yuanzhong and Chen, Dehao and Firat, Orhan and Huang, Yanping and Krikun, Maxim and Shazeer, Noam and Wu, Yonghui},
  booktitle={Proceedings of the International Conference on Learning Representations},
  year={2021}
}

@article{Krajewski2024ScalingLaws,
  title={Scaling Laws for Fine-Grained Mixture of Experts},
  author={Jakub Krajewski and Jan Ludziejewski and Kamil Adamczewski and Maciej Pióro
        and Michał Krutul and Szymon Antoniak and Kamil Ciebiera and Krystian Król
        and Tomasz Odrzygóźdź and Piotr Sankowski and Marek Cygan and Sebastian Jaszczur},
  journal={arXiv preprint arXiv:2402.07871},
  year={2024}
}

@article{Dadoun2025JacobianStability,
  title={On the Stability of the Jacobian Matrix in Deep Neural Networks},
  author={Dadoun, Benjamin and Hayou, Soufiane and Salam, Hanan and Seddik, Mohamed El Amine and Youssef, Pierre},
  journal={arXiv preprint arXiv:2506.08764},
  year={2025}
}

@article{Li2025SubMoE,
  title={Sub-MoE: Efficient Mixture-of-Expert LLMs Compression via Subspace Expert Merging},
  author={Li, Lujun and Zhu, Qiyuan and Wang, Jiacheng and Li, Wei and Gu, Hao and Han, Sirui and Guo, Yike},
  journal={arXiv preprint arXiv:2506.23266},
  year={2025}
}

@article{Miao2025MergeMoE,
  title={MergeMoE: Efficient Compression of MoE Models via Expert Output Merging},
  author={Miao, Ruijie and Yao, Yilun and Wang, Zihan and Wang, Zhiming and Yi, Bairen and Liu, LingJun and Zhao, Yikai and Yang, Tong},
  journal={arXiv preprint arXiv:2510.14436},
  year={2025}
}

@article{Yang2024MoEI2,
  title={MoE-I\textsuperscript{2}: Compressing Mixture of Experts Models through Inter-Expert Pruning and Intra-Expert Low-Rank Decomposition},
  author={Yang, Cheng and Sui, Yang and Xiao, Jinqi and Huang, Lingyi and Gong, Yu and Duan, Yuanlin and Jia, Wenqi and Yin, Miao and Cheng, Yu and Yuan, Bo},
  journal={arXiv preprint arXiv:2411.01016},
  year={2024}
}

@article{Huang2025MixtureCompressor,
  title={Mixture Compressor for Mixture-of-Experts LLMs Gains More},
  author={Huang, Wei and Liao, Yue and Liu, Jianhui and He, Ruifei and Tan, Haoru and Zhang, Shiming and Li, Hongsheng and Liu, Si and Qi, Xiaojuan},
  journal={arXiv preprint arXiv:2410.06270},
  year={2025}
}

@article{Edelman1998Orthogonality,
  title={The Geometry of Algorithms with Orthogonality Constraints},
  author={Edelman, Alan and Arias, Tom{\'a}s A. and Smith, Steven T.},
  journal={SIAM Journal on Matrix Analysis and Applications},
  year={1998}
}

@book{Absil2008Manifolds,
  title={Optimization Algorithms on Matrix Manifolds},
  author={Absil, P.-A. and Mahony, Robert and Sepulchre, Rodolphe},
  publisher={Princeton University Press},
  year={2008}
}

@book{Golub2013MatrixComputations,
  title={Matrix Computations},
  author={Golub, Gene H. and Van Loan, Charles F.},
  publisher={Johns Hopkins University Press},
  year={2013}
}

@article{Jacobs1991AdaptiveMixtures,
  title={Adaptive Mixtures of Local Experts},
  author={Jacobs, Robert A. and Jordan, Michael I. and Nowlan, Steven J. and Hinton, Geoffrey E.},
  journal={Neural Computation},
  volume={3},
  number={1},
  pages={79--87},
  year={1991}
}

@article{Krishnamurthy2023ImprovingExperts,
  title={Improving Expert Specialization in Mixture of Experts},
  author={Krishnamurthy, Yamuna and Watkins, Chris and Gaertner, Thomas},
  journal={arXiv preprint arXiv:2302.14703},
  year={2023}
}

@article{Eo2025MoCE,
  title={Mixture-of-Clustered-Experts: Advancing Expert Specialization and Generalization in Instruction Tuning},
  author={Eo, Sugyeong and Lee, Jungjun and Park, Chanjun and Lim, Heuiseok},
  journal={arXiv preprint arXiv:2509.10513},
  year={2025}
}

@article{Zoph2022STMoE,
  title={ST-MoE: Designing Stable and Transferable Sparse Expert Models},
  author={Zoph, Barret and Bello, Irwan and Kumar, Sameer and Du, Nan and Huang, Yanping and Dean, Jeff and Shazeer, Noam and Fedus, William},
  journal={arXiv preprint arXiv:2202.08906},
  year={2022}
}

@inproceedings{Jordan1994HME,
  title={Hierarchical Mixtures of Experts and the EM Algorithm},
  author={Jordan, Michael I. and Jacobs, Robert A.},
  booktitle={Proceedings of the International Joint Conference on Neural Networks},
  year={1994}
}

@article{Dai2024DeepSeekMoE,
  title={DeepSeekMoE: Towards Ultimate Expert Specialization in Mixture-of-Experts Language Models},
  author={Dai, Damai and Deng, Chengqi and Zhao, Chenggang and Xu, R.X. and Gao, Huazuo and Chen, Deli and Li, Jiashi and Zeng, Wangding and Yu, Xingkai and Wu, Y. and Xie, Zhenda and Li, Y.K. and Huang, Panpan and Luo, Fuli and Ruan, Chong and Sui, Zhifang and Liang, Wenfeng},
  journal={arXiv preprint arXiv:2401.06066},
  year={2024}
}

@article{Novak2018Sensitivity,
  title={Sensitivity and Generalization in Neural Networks: An Empirical Study},
  author={Novak, Roman and Bahri, Yasaman and Abolafia, Daniel A. and Pennington, Jeffrey and Sohl-Dickstein, Jascha},
  journal={arXiv preprint arXiv:1802.08760},
  year={2018}
}

@inproceedings{Pennington2018Spectral,
  title={The Emergence of Spectral Universality in Deep Networks},
  author={Pennington, Jeffrey and Schoenholz, Samuel S. and Ganguli, Surya},
  booktitle={Proceedings of the 21st International Conference on Artificial Intelligence and Statistics},
  year={2018}
}

@inproceedings{Ethayarajh2019Contextual,
  title={How Contextual Are Contextualized Word Representations? Comparing the Geometry of BERT, ELMo, and GPT-2 Embeddings},
  author={Ethayarajh, Kawin},
  booktitle={Proceedings of EMNLP},
  year={2019}
}

@inproceedings{Timkey2021RogueDimensions,
  title={All Bark and No Bite: Rogue Dimensions in Transformer Language Models Obscure Representational Quality},
  author={Timkey, William and van Schijndel, Marten},
  booktitle={Proceedings of EMNLP},
  year={2021}
}

\appendix

\section{Additional notation, definitions, and methodological details}
\subsection{MoE Layer and Routing Functions}
For an MoE layer with routing weights $g(x) \in \mathbb{R}^E$ and a collection of $E$ experts $\{ f_e \}_{e=1}^E$,
For an input hidden state $x \in \mathbb{R}^d$, the layer output is

\begin{equation}
f_{\text{MoE}}(x) = \sum_{e=1}^{E} g_e(x)\, f_e(x).
\end{equation}

\textbf{Top-k routing:} 
The router produces a sparse vector $g(x)$, let $\operatorname{TopK}(x) \subseteq \{1, \ldots, E\}$ denotes the selected experts, then

\begin{equation}
g_e(x) = 
\begin{cases} 
\dfrac{\exp(s_e(x))}{\sum_{e' \in \operatorname{TopK}(x)} \exp(s_{e'}(x))}, & e \in \operatorname{TopK}(x) \\[10pt]
0, & \text{otherwise}
\end{cases}
\end{equation}

Where $s_e(x)$ is the router logits.

\textbf{Fully-soft routing:}
All experts receive non-zero weight:
\begin{equation}
g_e(x) = \frac{\exp(s_e(x))}{\sum_{j=1}^E \exp(s_j(x))}
\end{equation}

\subsection{The Step-by-Step Calculation on Grassmann distance}
Let $Q_i,\, Q_j \in \mathbb{R}^{d \times n}$ denote the top-$n$ PCA principal subspaces for experts $e_i$ and $e_j$, 
the Grassmannian Geodesic Distance is computed as follows:

\paragraph{1. Projection Overlap.}
Multiply the transposed subspace of one expert with the subspace of the other to obtain the overlap matrix:
\begin{equation}
M = Q_1^\top Q_2 .
\end{equation}
This results in a $n \times n$ matrix.

\paragraph{2. Singular Value Decomposition of the Overlap Matrix.}
Perform SVD on the overlap matrix:

\begin{equation}
M = U \Sigma V^\top ,
\end{equation}

where the singular values satisfy $\sigma_m = \cos(\theta_m), m = 1,\ldots,n.$ and $\theta_m$ are the 
principal angles between the subspaces spanned by $Q_1$ and $Q_2$.

\paragraph{3. Principal Angles.}
Extract the principal angles via
\begin{equation}
\theta_m = \arccos(\sigma_m).
\end{equation}

\paragraph{4. Grassmann Geodesic Distance.}
The canonical Grassmannian distance is the \(\ell_2\) norm of the principal angles:

\begin{equation}
d_G(e_i, e_j) = \left( \sum_{m=1}^{n} \theta_m^2 \right)^{1/2}.
\end{equation}

Where $\theta_m = \arccos(\sigma_m) \in [0, \tfrac{\pi}{2}]$, and $\sigma_1,\, \sigma_2,\, \ldots,\, \sigma_n$ are singular values 
of $M = Q_i^{\top} Q_j \in \mathbb{R}^{n \times n}$.  
Theoretically, the maximum distance when perfectly orthogonal is $\frac{\pi}{2}\,\sqrt{n}$. 
For $n=5$, that maximum is approx 3.51. 

\section{Experimental Setup for Pretrained Models}
\subsection{Models}
\begin{itemize}
    \item \textbf{Mistral-7B (dense) vs Mistral-8×7B (MoE):} 
    We analyze the Mistral family of models, comparing the dense 7B parameter variant with the MoE 8×7B variant. 
    Both models share a similar architecture and training regime, allowing us to isolate the effect of MoE routing on geometric structure.

    \item \textbf{Qwen2.5-1.5B (dense) vs Qwen1.5-MoE-A2.7B (MoE):} 
    We also analyze the Qwen family, comparing the dense 1.5B parameter variant with the MoE A2.7B variant. 
    This provides a complementary test of whether the geometric patterns observed in Mistral generalize to a different architecture and scale.
\end{itemize}

\textbf{Expert configurations for the MoE models} are as follows:
\begin{itemize}
    \item \textbf{Mistral-8×7B:} This model contains 8 experts per MoE layer, with each expert having 7B parameters. 
    The routing mechanism is Top-k, where the router selects the top 2 experts for each input token.

    \item \textbf{Qwen1.5-MoE-A2.7B:} This model contains 60 experts per MoE layer, with each expert having approximately 45M parameters (totaling 2.7B parameters for the MoE layers). 
    The routing mechanism is also Top-k, with the router selecting the top 4 experts for each input token.
\end{itemize}   

\subsection{Data}
We use two distinct text corpora, each exceeding 16k tokens, for our analysis. 
The batch size is set to 512 tokens.

\section{Experimental Setup for Controlled Routing Experiments}
\subsection{Models}
We evaluate:
\begin{itemize}
       \item \textbf{MoE-Top-k.} A sparsely-activated MoE layer using Top-k routing with k=2 unless otherwise specified.

    \item \textbf{MoE-Fully-Soft.} A fully-soft MoE layer using a softmax router over all experts.
\end{itemize}
All models share the same hidden dimension, number of heads, feedforward width, and activation functions. 
The only difference across conditions is the routing mechanism.

\textbf{Expert configuration:}
\begin{itemize}
    \item Number of layers: 3
    \item Number of experts: $E=8$ 
    \item Expert architecture: 2-layer MLP with GELU
    \item Model dimension: $d_model=128$
    \item Expert hidden dimension (same as dense FFN): $d_hidden=256$
    \item Block size: 64
    \item Batch size: 32
    \item Router: single linear projection followed by softmax or Top-k selection
\end{itemize}

This configuration provides a balanced regime where 
(i) routing behavior is non-trivial, 
(ii) Jacobian computation remains tractable, and 
(iii) PCA spectra are stable across seeds. 
The dimensions are chosen to match the per‑expert capacity of small transformer 
feedforward blocks while enabling full spectral analysis without approximation.

\subsection{Routing Settings}
\textbf{Top-k routing}

For the Top-k MoE configuration, the router produces a sparse distribution over experts:
\begin{itemize}
    \item Top-k: k=2
    \item Routing weights: obtained by applying a softmax to router logits, then selecting the top-k entries
    \item Temperature: 1.0
    \item Normalization: selected expert weights are renormalized to sum to 1
\end{itemize}
This setting enforces sharp expert selection and produces sparse expert activation patterns.

\textbf{Fully-soft routing}

For the fully-soft MoE configuration, the router assigns non-zero weight to all experts:
\begin{itemize}
    \item Routing weights: full softmax over all E experts
    \item No sparsity constraints: every expert contributes to the output
    \item Temperature: 1.0
    \item No renormalization: softmax already produces a valid probability distribution
\end{itemize}
This setting yields diffuse expert activation and higher-entropy routing behavior.

\textbf{Batching and weighting}

Routing weights $g_e(x)$ are computed per token (per sample in the synthetic batch). 
These weights are used consistently across all probes:
\begin{itemize}
    \item PCA on routing-weighted hidden states: routing weights serve as sample weights for expert‑specific covariance estimation
    \item Jacobian averaging: expert‑local Jacobians are averaged using the same per‑token routing weights
\end{itemize}
This ensures that each expert’s spectral geometry reflects the distribution of inputs it actually receives under the router.

\subsection{Training Details}
Both MoE models are trained using cross entropy loss on tiny wikitext data. We use the Adam optimizer with a learning rate of $3 \times 10^{-4}$. 
Each model is trained independently using the same batch of data to ensure that differences in spectral geometry arise solely 
from the routing mechanism.

\subsection{Hardware}
All experiments are executed on an Apple M1 Pro system (MacBook Pro, 16 GB RAM). Computations are performed using the default 
framework backend without explicit configuration for GPU/Metal (MPS) acceleration. The small-scale 3-layer transformer 
and WikiText setup make full Jacobian computation tractable under this regime, while avoiding reliance 
on hardware-specific optimizations.

\section{Additional Results: Qwen Geometric Structure}
\label{app:qwen}

\begin{figure}[t]
    \centering

    \begin{subfigure}{0.48\linewidth}
        \centering
        \includegraphics[width=\linewidth]{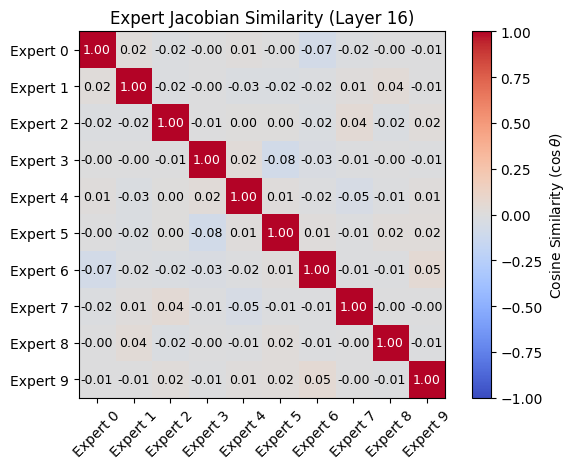}
        \caption{}
    \end{subfigure}
    \hfill
    \begin{subfigure}{0.48\linewidth}
        \centering
        \includegraphics[width=\linewidth]{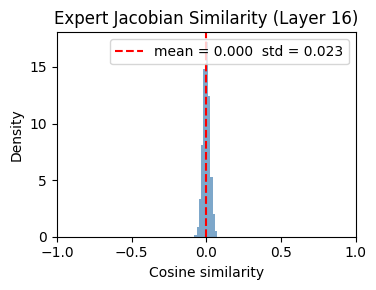}
        \caption{}
    \end{subfigure}

    \vspace{0.25em}

    \begin{subfigure}{0.48\linewidth}
        \centering
        \includegraphics[width=\linewidth]{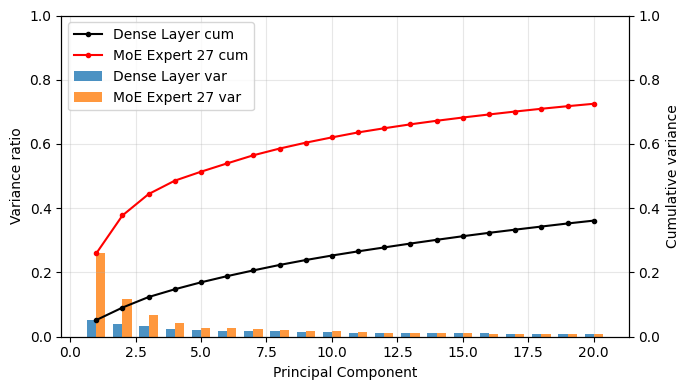}
        \caption{}
    \end{subfigure}
    \hfill
    \begin{subfigure}{0.48\linewidth}
        \centering
        \includegraphics[width=\linewidth]{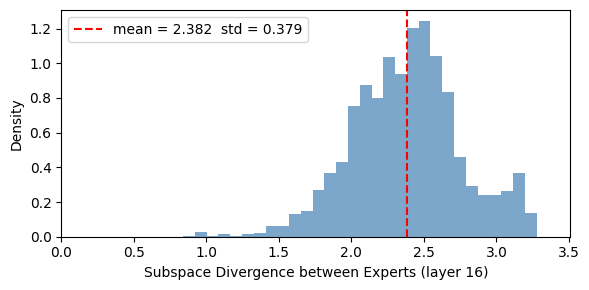}
        \caption{}
    \end{subfigure}

    \caption{\textbf{Geometric structure of MoE specialization in Qwen1.5‑MoE‑A2.7B (Layer 16).}
    (a) Cross-expert Jacobian similarity matrix (first 10 experts shown).
    (b) Distribution of Jacobian similarities.
    (c) Routed PCA spectra for dense vs.\ MoE expert layer.
    (d) Grassmannian distances between expert subspaces (top-5 components; theoretical maximum $\approx 3.51$).}

\end{figure}

\section{Additional Results: Controlled Routing Experiments}
\label{app:controlled}
\begin{figure}[t]
    \centering

    \begin{subfigure}{0.48\linewidth}
        \centering
        \includegraphics[width=\linewidth]{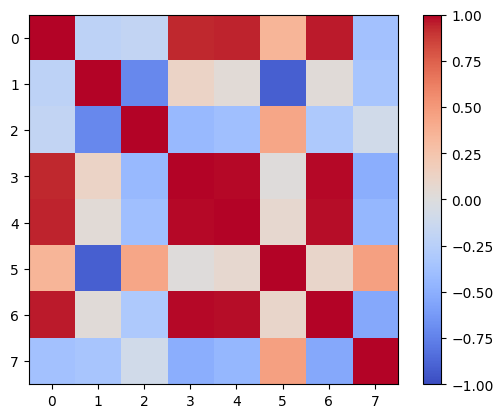}
        \caption{}
    \end{subfigure}
    \hfill
    \begin{subfigure}{0.48\linewidth}
        \centering
        \includegraphics[width=\linewidth]{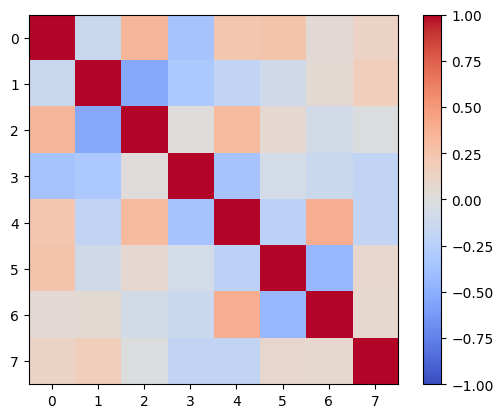}
        \caption{}
    \end{subfigure}

    \vspace{0.25em}

    \begin{subfigure}{0.48\linewidth}
        \centering
        \includegraphics[width=\linewidth]{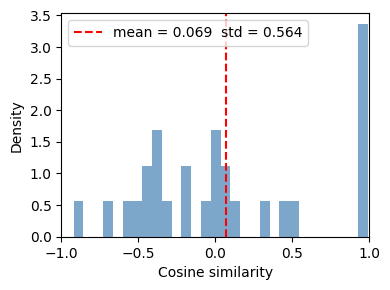}
        \caption{}
    \end{subfigure}
    \hfill
    \begin{subfigure}{0.48\linewidth}
        \centering
        \includegraphics[width=\linewidth]{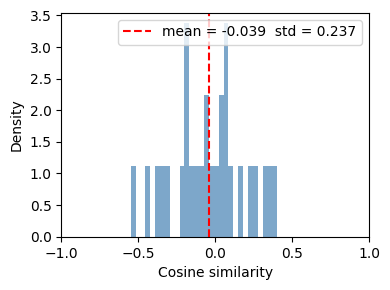}
        \caption{}
    \end{subfigure}

    \vspace{0.25em}

    \caption{\textbf{ Effect of routing sharpness on MoE geometry in the controlled 3-layer Transformer model.}
    (a-b) Jacobian similarity matrix under fully-soft vs Top-k routing. 
    (c-d) Distribution of Jacobian similarities under fully-soft vs Top-k routing.}

\end{figure}

\begin{figure}[t]
    \centering

    \begin{subfigure}{0.48\linewidth}
        \centering
        \includegraphics[width=\linewidth]{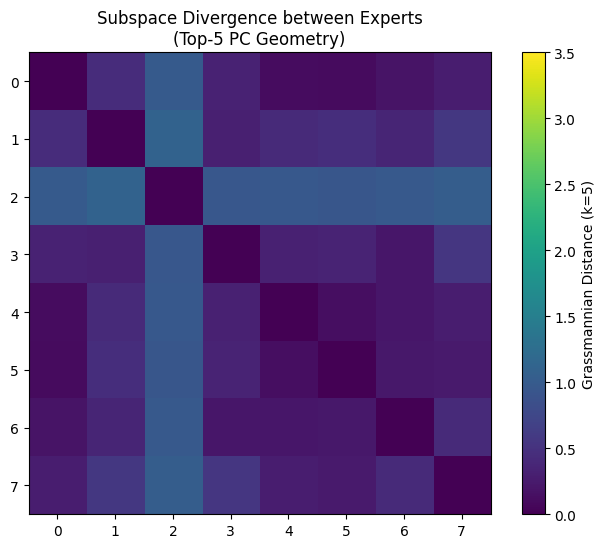}
        \caption{}
    \end{subfigure}
    \hfill
    \begin{subfigure}{0.48\linewidth}
        \centering
        \includegraphics[width=\linewidth]{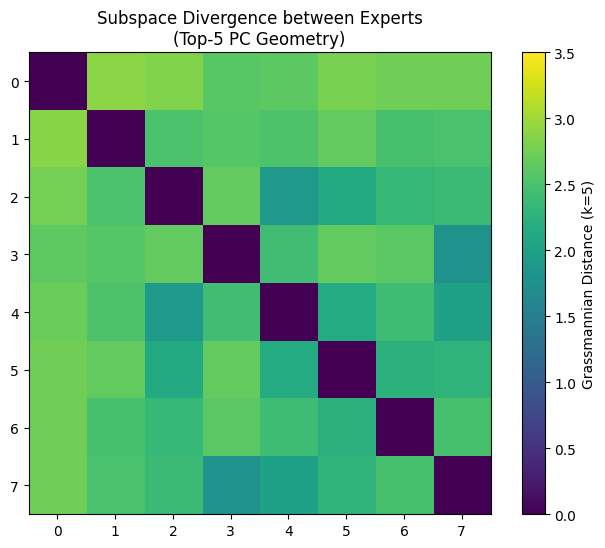}
        \caption{}
    \end{subfigure}

    \vspace{0.25em}

    \begin{subfigure}{0.48\linewidth}
        \centering
        \includegraphics[width=\linewidth]{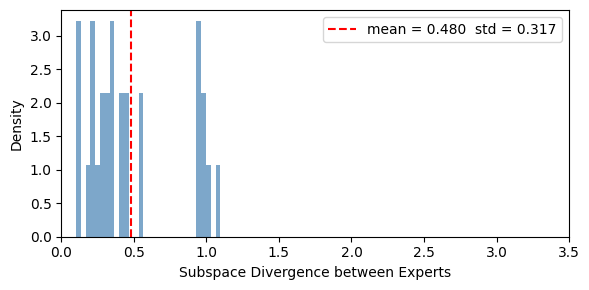}
        \caption{}
    \end{subfigure}
    \hfill
    \begin{subfigure}{0.48\linewidth}
        \centering
        \includegraphics[width=\linewidth]{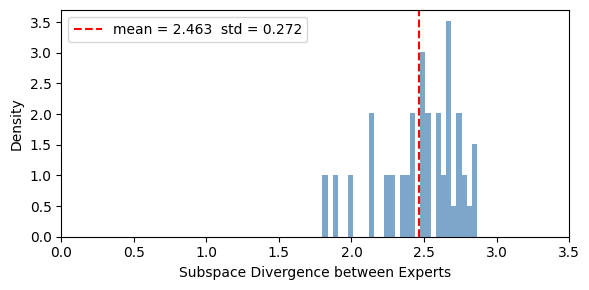}
        \caption{}
    \end{subfigure}

    \caption{\textbf{Effect of routing sharpness on MoE geometry in the controlled 3-layer Transformer model.}
    (a–b) Grassmannian distance matrix under fully-soft vs.\ Top-k routing.
    (c–d) Distribution of Grassmannian distances under fully-soft vs.\ Top-k routing (theoretical maximum $\approx 3.51$).}

\end{figure}

\end{document}